\documentclass[pdflatex,sn-mathphys-num]{sn-jnl}

\usepackage{graphicx}%
\usepackage{multirow}%
\usepackage{amsmath,amssymb,amsfonts}%
\usepackage{amsthm}%
\usepackage{mathrsfs}%
\usepackage[title]{appendix}%
\usepackage{xcolor}%
\usepackage{textcomp}%
\usepackage{manyfoot}%
\usepackage{booktabs}%
\usepackage{algorithm}%
\usepackage{algorithmicx}%
\usepackage{algpseudocode}%
\usepackage{listings}%

\usepackage{siunitx}
\usepackage{float}
\usepackage{threeparttable}

\usepackage{graphicx}

\usepackage{tabularray}
\UseTblrLibrary{booktabs}  

\usepackage{booktabs}

\usepackage[normalem]{ulem}

\usepackage{rotating}

\usepackage{float}    
\usepackage{booktabs} 

\theoremstyle{thmstyleone}%
\theoremstyle{thmstyletwo}%

\theoremstyle{thmstylethree}%

\begin{document}

\title[Article Title]{Enhancing Spatio-Temporal Forecasting with Spatial Neighbourhood Fusion: A Case Study on Mobility in Peru}


\author*[1,2,3,4]{\fnm{Chuan} \sur{Li}}\email{chuan.li@sorbonne-universite.fr}

\author[5,6]{\fnm{Jiang} \sur{You}}\email{jiang.you@esiee.fr}

\author[1,3,4]{\fnm{Hassine} \sur{Moungla}}\email{hassine.moungla@parisdescartes.fr}

\author[7,4]{\fnm{Vincent} \sur{Gauthier}}\email{vincent.gauthier@telecom-sudparis.eu}

\author[8]{\fnm{Miguel} \sur{Nunez-del-Prado}}\email{mnunezdelpradoco@worldbank.org}

\author[9]{\fnm{Hugo} \sur{Alatrista-Salas}}\email{hugo.alatrista\_salas@devinci.fr}

\affil*[1]{\orgname{LIPADE, Université Paris Cité},
\orgaddress{\city{Paris}, \country{France}}}

\affil[2]{\orgname{EDITE, Sorbonne Université},
\orgaddress{\city{Paris}, \country{France}}}

\affil[3]{\orgname{Télécom SudParis},
\orgaddress{\city{Palaiseau}, \country{France}}}

\affil[4]{\orgname{Institut Polytechnique de Paris},
\orgaddress{\city{Palaiseau}, \country{France}}}

\affil[5]{\orgname{LISSI, Université Paris-Est Créteil},
\orgaddress{\city{Paris}, \country{France}}}

\affil[6]{\orgname{ESIEE Paris - Université Gustave Eiffel (UGE)},
\orgaddress{\city{Paris}, \country{France}}}

\affil[7]{\orgname{SAMOVAR, Télécom SudParis},
\orgaddress{\city{Palaiseau}, \country{France}}}

\affil[8]{\orgname{The World Bank},
\orgaddress{\city{Washington, DC}, \country{USA}}}

\affil[9]{\orgname{De Vinci Higher Education, De Vinci Research Center},
\orgaddress{\city{Paris}, \country{France}}}


\abstract{Accurate modeling of human mobility is critical for understanding epidemic spread and deploying timely interventions. In this work, we leverage a large-scale spatio-temporal dataset collected from Peru's national Digital Contact Tracing (DCT) application during the COVID-19 pandemic to forecast mobility flows across urban regions. A key challenge lies in the spatial sparsity of hourly mobility counts across hexagonal grid cells, which limits the predictive power of conventional time series models. To address this, we propose a lightweight and model-agnostic \textit{Spatial Neighbourhood Fusion} (SPN) technique that augments each cell’s features with aggregated signals from its immediate H3 neighbors. We evaluate this strategy on three forecasting backbones—NLinear, GRU, LSTM, PatchTST, and K-U-Net—under various historical input lengths. Experimental results show that SPN consistently improves forecasting performance, achieving up to 9.92\% reduction in test MSE(Mean Square Error). Our findings demonstrate that spatial smoothing of sparse mobility signals provides a simple yet effective path toward robust spatio-temporal forecasting during public health crises.}




\maketitle

\section{Introduction}
Human mobility is one of the strongest macroscopic drivers of infectious-disease dynamics.  
During the COVID-19 pandemic, aggregated mobile-phone traces provided unprecedented visibility into how non-pharmaceutical interventions (NPIs) reshaped population movement and, in turn, epidemic curves \cite{oliver2020mobile,sciadv2021mobility}. 
Digital contact-tracing (DCT) platforms push this visibility to an even finer scale, recording person-to-person proximity events in near real time.  
Peru was an early adopter: its Perú en tus manos app merged Bluetooth encounters with DCT positioning and has remained one of the few fully operational national DCT systems in Latin America \cite{peru_dct2021,li2025dct,li2025assessing}.
The resulting data stream, covering millions of users, opens a unique opportunity to model urban mobility flows in settings where official traffic sensors or ticketing logs are scarce.

Yet transforming raw proximity pings into reliable origin–destination (OD) counts is non-trivial.  Spatial discretisation with hexagonal H3 cells\cite{h3_geo} offers a principled way to regularise geometry and enable efficient spatial joins \cite{uber_h3}. Fine-grained grid-based representations have also been explored for clustering and urban-scale pattern mining from mobile metadata \cite{hu2023spatio, hu2025fine}. 
At the hourly resolution required for real-time public-health response, however, most cells contain few or zero trips, leading to extreme sparsity \cite{dglm_sparse_mobility2024}.
While deep spatio-temporal models—ranging from graph neural networks to Transformer variants—can learn complex dependencies, they often overfit noisy zero-inflated inputs or fail to extrapolate when data gaps appear \cite{histgnn2022,Nie-2023-PatchTST}.

We address this challenge with a lightweight Spatial Neighbourhood Fusion (SPN) scheme that augments each H3 cell’s signal with robust statistics from its immediate neighbours.  SPN acts as a plug-and-play preprocessing layer and can be paired with any forecasting backbone.  Using a large-scale dataset from Peru’s national DCT deployment, we show that SPN consistently reduces test-set MSE across five representative models (NLinear,GRU, LSTM, PatchTST, K-U-Net) and four input horizons, highlighting the value of local spatial smoothing for epidemic-era mobility prediction.

\section{Related Work}
Early COVID-19 studies established that decreases in inter-provincial travel preceded lower case growth in multiple regions. Oliver et al. synthesised best practices for using telecom data throughout the pandemic life-cycle, emphasising privacy safeguards and representativeness \cite{oliver2020mobile}.
Parallel work coupled mobility networks with compartmental models or gravity kernels to explain city-level reproduction numbers \cite{gravity_covid2025}.

Beyond aggregated tower pings, DCT platforms offer metre-level resolution of interpersonal encounters.  
Serafino et al. demonstrated that billions of GPS-level contacts capture the changing topology of transmission networks and can inform targeted quarantine policies \cite{serafino2022dct}.
Peru’s Perú en tus manos remains a flagship example of nationwide deployment, providing anonymised GPS + Bluetooth traces that have been used for both epidemiological analysis and mobility research \cite{peru_dct2021,li2025dct,li2025assessing}.
Graph-based deep learners such as HiSTGNN \cite{histgnn2022} and ST-GNN variants \cite{jin2023spatio} fuse spatial adjacency with temporal convolutions to predict case counts or traffic volumes, but require dense observations at every node.  
Transformer architectures have been adapted to long-horizon time-series tasks; PatchTST segments inputs into patches to alleviate quadratic attention costs and currently leads many public benchmarks \cite{Nie-2023-PatchTST}.
Conversely, the NLinear family of embarrassingly simple linear nets questions whether heavy architectures are needed at all \cite{Zeng_AreTE_2022}.
All these models, however, are sensitive to the zero-inflation that arises when fine-grid mobility is sparsely sampled.
Neighbourhood aggregation is a classical remedy for sparse geospatial counts, from remote-sensing gap-filling \cite{spatiotemporal_smoothing2021} to adaptive traffic forecasting \cite{adstgcn2023}. 
Median-based fusion, in particular, is robust to outliers and has low computational overhead—properties that motivate our SPN design.

\smallskip
\noindent\textbf{Positioning of this work.}  
Where prior studies rely on complex graph diffusion or attention layers to propagate information, we show that a model-agnostic, first-order neighbourhood median is sufficient to boost performance on a real, country-scale DCT dataset.  SPN thus complements, rather than competes with, existing deep architectures and can be deployed at almost no extra cost, a feature that is crucial for low-resource public-health settings.

\section{Data Pre-processing}

We analyse GPS traces collected by Peru’s national Digital Contact Tracing (DCT) application between April and June 2020.  
The raw dataset contains \SI{808}{M} location pings from \SI{1.58}{M} users.  
Restricting the sample to “active” users—those with at least 30 pings—yields \SI{651}{k} reliable trajectories.  
Descriptive statistics appear in Table~\ref{tab:summary-stats}. Pre-processin by Algorithm \ref{alg:pipeline}.


\begin{table}[!htbp]
\caption{Dataset overview}
\label{tab:summary-stats}
\centering
\begin{tabular}{@{}p{4.9cm}r@{}}
\toprule
Records with DCT fix                                        & 808{,}021{,}726 \\
Users with DCT records                                      & 1{,}581{,}867 \\
Active users ($\ge$30 records)                              & 651{,}155  \\
App downloads (census)                                      & 1{,}896{,}228 \\
Study period                                                & Apr.--Jun.\ 2020 \\
Cumulative infected users                                   & 81{,}305 \\
Infected users with DCT data                                & 10{,}131 \\
\bottomrule
\end{tabular}
\end{table}

\begin{algorithm}[t]
\caption{Pre-processing pipeline for DCT mobility data}
\label{alg:pipeline}
\begin{algorithmic}[1]
\Require Raw records $(u,t,\varphi,\lambda,\textit{status})$
\Ensure Hourly tensors $\mathsf{OD}$, $\mathsf{IN}$, $\mathsf{OUT}$
\State $A \gets \{u:\ \#\text{pings}(u)\ge 30\}$ \Comment{active users}
\State $\mathcal{R} \gets \{r\in \text{records}:\ u(r)\in A,\ t(r)\in \text{Apr.--Jun.\ 2020}\}$
\State $\mathcal{S} \gets \emptyset$ \Comment{stop points}

\ForAll{$u\in A$}
  \State $\mathcal{C} \gets \emptyset$ \Comment{current cluster}
  \ForAll{points $p$ of user $u$ in chronological order}
    \If{$\operatorname{dist}\!\bigl(p,\operatorname{medoid}(\mathcal{C})\bigr)\le \SI{50}{m} \ \land\ \Delta t \le \SI{5}{min}$}
      \State $\mathcal{C} \gets \mathcal{C}\cup\{p\}$
    \Else
      \If{$\operatorname{duration}(\mathcal{C}) \ge \SI{5}{min}$}
        \State $\mathcal{S} \gets \mathcal{S}\cup\{\operatorname{medoid}(\mathcal{C})\}$
      \EndIf
      \State $\mathcal{C} \gets \{p\}$
    \EndIf
  \EndFor
\EndFor

\ForAll{$\sigma\in\mathcal{S}$}
  \State $\sigma.\textit{h3} \gets \textsc{GeoToH3}(\sigma.\varphi,\sigma.\lambda,6)$
\EndFor

\State Initialise $\mathsf{OD}[d,h,s,t] \gets 0$
\ForAll{$u\in A$}
  \ForAll{consecutive stops $(\sigma_k,\sigma_{k+1})$ of user $u$}
    \State $(d,h) \gets$ (date, hour) of $\sigma_k$
    \State $\mathsf{OD}[d,h,\sigma_k.\textit{h3},\sigma_{k+1}.\textit{h3}] \gets \mathsf{OD}[d,h,\sigma_k.\textit{h3},\sigma_{k+1}.\textit{h3}] + 1$
  \EndFor
\EndFor

\ForAll{$(d,h,c)$}
  \State $\mathsf{IN}[d,h,c] \gets \sum_{s}\mathsf{OD}[d,h,s,c]$
  \State $\mathsf{OUT}[d,h,c] \gets \sum_{t}\mathsf{OD}[d,h,c,t]$
\EndFor
\State \Return $\mathsf{OD}$, $\mathsf{IN}$, $\mathsf{OUT}$
\end{algorithmic}
\end{algorithm}

\subsection{Spatial Neighbourhood Feature Fusion (SPN)}

Due to the sparsity of mobility signals in certain \texttt{H3} cells at specific timestamps, we propose a lightweight yet effective technique named \textbf{Spatial Neighbourhood Fusion (SPN)} to enhance temporal continuity through local spatial aggregation.

Given a hexagonal cell \( c \) at time \( t \), we define its estimated spatial flow as the median of non-zero total flows among its immediate neighbors:
\begin{align}
\hat{f}_{c}^{(t)} = \operatorname*{median} \left\{ f_{n}^{(t)} \;\middle|\; n \in \mathcal{N}(c),\ f_{n}^{(t)} > 0 \right\},
\end{align}
where \( f_{n}^{(t)} \) denotes the observed total flow in cell \( n \) at time \( t \), and \( \mathcal{N}(c) = \texttt{grid\_disk}(c,1) \setminus \{c\} \) is the first-order spatial neighborhood defined by the \texttt{H3} indexing system.

If no valid neighbors exist (i.e., all neighbors are missing or zero-valued), we fallback to the H3 cell's original value: \( \hat{f}_{c}^{(t)} \leftarrow f_{c}^{(t)} \).

To further stabilize the signal, we compute the arithmetic mean of the original and spatially aggregated flow:
\begin{align}
\bar{f}_{c}^{(t)} = \frac{1}{2} \left(f_{c}^{(t)} + \hat{f}_{c}^{(t)}\right).
\end{align}

Finally, we construct a three-channel feature vector for each cell and time step as:
\[
\mathbf{x}_{c}^{(t)} = \left[f_{c}^{(t)},\; \hat{f}_{c}^{(t)},\; \bar{f}_{c}^{(t)}\right].
\]

This SPN operation is non-parametric, model-agnostic, and computationally efficient (under \( 4\,\mathrm{ms} \) per hourly snapshot on standard hardware).

\subsection{Night-lights-guided Urban Mask}
\begin{figure}[!t]
  \centering
  \includegraphics[width=\linewidth]{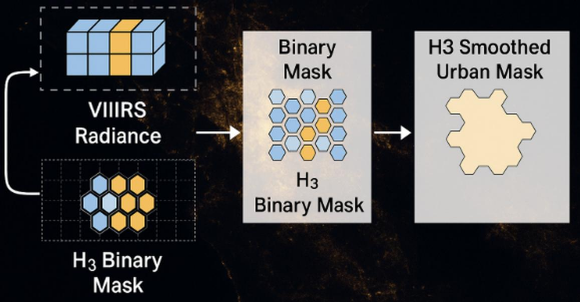}
  \caption{Urban mask construction guided by night-time lights.
  Raw VIIRS radiance is first aligned with the H3 grid system, yielding a binary
  mask where cells above the radiance threshold (\SI{8}{nW\,cm^{-2}\,sr^{-1}}) are marked as urban.
  This binary urban mask is then smoothed using kernel density estimation to form
  contiguous urban clusters. The resulting H3-smoothed mask enables downstream models
  to differentiate between urban and peri-urban dynamics.}
  \label{fig:urban-mask}
\end{figure}
To distinguish urban from peri-urban H3 cells, we leverage radiance data from the Visible Infrared Imaging Radiometer Suite (VIIRS) \cite{elvidge2017viirs}.
Empirical studies show that urban cores typically exceed \SI{10}{nW\,cm^{-2}\,sr^{-1}} \cite{small2021spatiotemporal}, where the unit \textit{nW\,cm$^{-2}$\,sr$^{-1}$} denotes nanowatts of radiative power per square centimeter per steradian—a standard measure of optical radiance. We adopt a conservative threshold of \SI{8}{nW\,cm^{-2}\,sr^{-1}}, then smooth the resulting binary mask using kernel density estimation to form contiguous urban clusters \cite{wang2021comparing}. Temporal stacks of VIIRS images allow us to track urban growth over the study window \cite{cao2022exploring}, providing an external validity check for the SPN-enhanced mobility flows.

\section{Method}
\begin{table}[t]
\caption{Multivariate time–series forecasting on the {\em smoothed} mobility–flow dataset. Look-back $L\!\in\!\{48,72,96,120\}$; horizon $T\!=\!48$.  Lower MSE is better.  Best scores are \textbf{bold}; second best are \uline{underlined}. “Avg Imp.” reports the \% improvement of each “+ spn’’ variant over its backbone (same smoothed target).}

\label{tab:multivariate-table}
\centering
\scriptsize
\resizebox{\textwidth}{!}{%
\begin{tblr}{
  colspec = {l c *{20}{c}},
  cell{1}{1} = {c=2}{},
  cell{2}{1} = {c=2}{},
  cell{7}{1} = {c=2}{},
  cell{1}{3} = {c=2}{},
  cell{1}{5} = {c=2}{},
  cell{1}{7} = {c=2}{},
  cell{1}{9} = {c=2}{},
  cell{1}{11} = {c=2}{},
  cell{1}{13} = {c=2}{},
  cell{1}{15} = {c=2}{},
  cell{1}{17} = {c=2}{},
  cell{1}{19} = {c=2}{},
  cell{1}{21} = {c=2}{},
  hline{1-8} = {-}{},
  vline{2-3,5,7,9,11,13,15,17,19,21} = {-}{},
}
Method & & Linear &       & Linear+spn &      & GRU &       & GRU+spn &      & LSTM &       & LSTM+spn &      & PatchTST &       & PatchTST+spn &      & K-U-Net &       & K-U-Net+spn & \\
Metric $(L)$ & & Train & Test & Train & Test & Train & Test & Train & Test & Train & Test & Train & Test & Train & Test & Train & Test & Train & Test & Train & Test \\
\begin{sideways}Length\end{sideways}
 & 48  & 0.4343 & 0.4146 & 0.4033 & 0.3834 & 0.2312 & 0.3927 & 0.2199 & \textbf{0.3382} & 0.2298 & 0.3959 & 0.2237 & 0.3682 & 0.4097 & 0.3848 & 0.3753 & 0.3675 & 0.3553 & 0.3838 & 0.3173 & \uline{0.3535} \\
 & 72  & 0.4160 & 0.3959 & 0.3828 &0.3615 & 0.2265 & 0.4065 & 0.2153 & \textbf{0.3377} & 0.2342 & 0.3853 & 0.2266 & 0.3758 & 0.3993 & 0.3780 & 0.3588 & 0.3493 & 0.3155 & 0.3758 & 0.3002 & \uline{0.3459} \\
 & 96  & 0.4138 & 0.3902 & 0.3798 & 0.3539 & 0.2170 & 0.4050 & 0.2090 & 0.3530 & 0.2300 & 0.3903 & 0.2176 & \textbf{0.3326} & 0.3950 & 0.3818 & 0.3565 & 0.3483 & 0.2963 & 0.3745 & 0.2718 & \uline{0.3462} \\
 & 120 & 0.4173 & 0.4098 & 0.3829 & 0.3742 & 0.2301 & 0.3894 & 0.2199 & 0.3305 & 0.2342 & 0.3804 & 0.2221 & \textbf{0.3229} & 0.3951 & 0.3826 & 0.3528 & \uline{0.3399} & 0.3218 & 0.3937 & 0.2717 & 0.3472 \\
Avg Imp. (\%) & & – & – & \textbf{46.13} & \textbf{50.49} & – & – & \textbf{4.50} & \textbf{14.70} & – & – & \textbf{4.11} & \textbf{9.82} & – & – & \textbf{9.74} & \textbf{8.00} & – & – & \textbf{9.92} & \textbf{8.84} \\
\bottomrule
\end{tblr}}
\end{table}

\begin{figure}[!t]
  \centering

  \includegraphics[width=\linewidth]{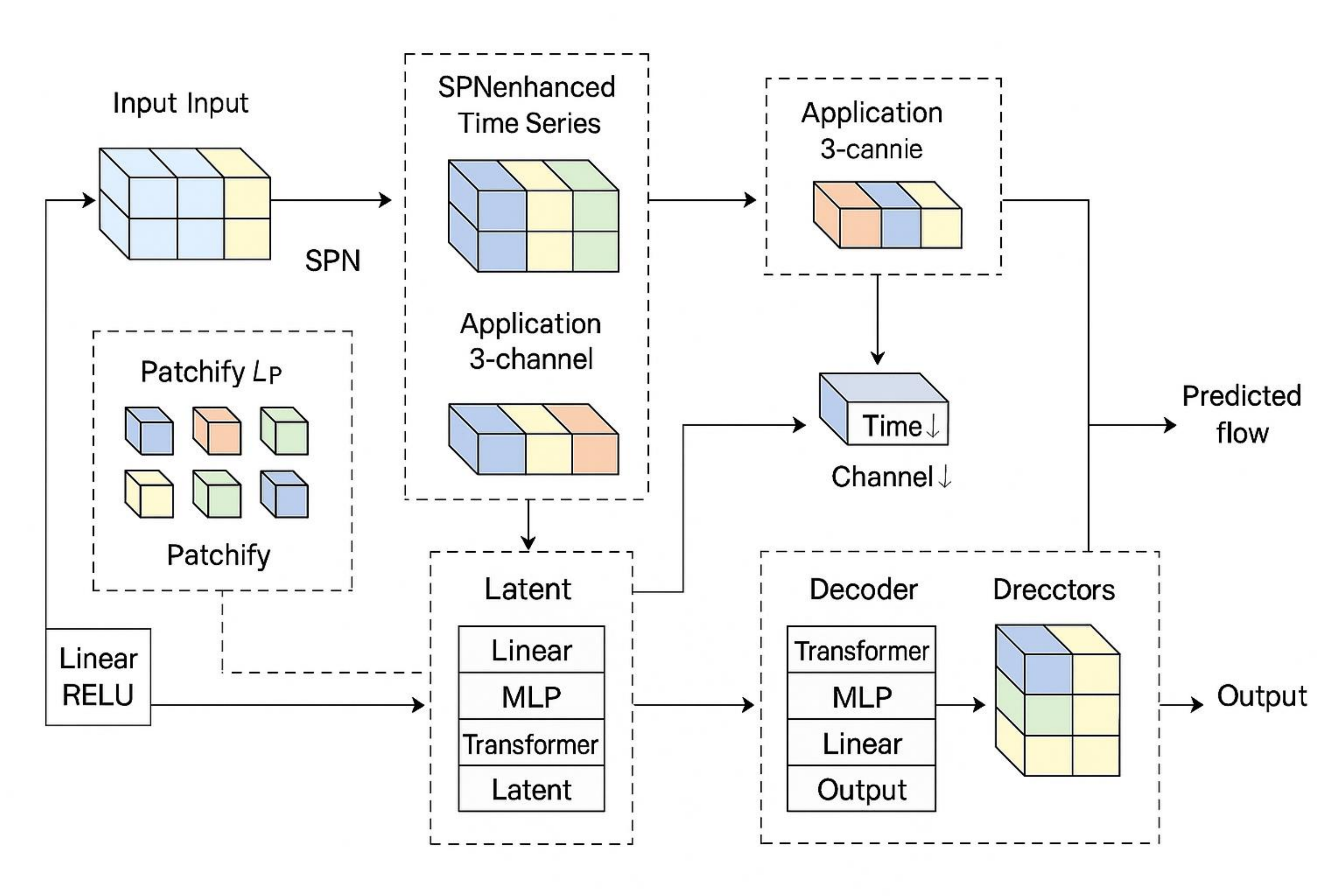}
  \vspace{-1em}
  \caption{End-to-end forecasting pipeline.
  An \textbf{SPN} block (left, not shown) fuses each H3 cell with the median of its
  six neighbours; the resulting three-channel tensor ($L\times3$) is patch-embedded
  and fed into one of three backbones.
  \textit{NLinear} applies channel-wise linear layers;
  \textit{PatchTST} operates on non-overlapping patches with a Transformer encoder;
  \textit{Kernel U-Net} (centre) uses a symmetric encoder–decoder whose blocks wrap
  custom kernels (Linear, MLP, LSTM, Transformer).
  Skip connections preserve high-resolution context, and the latent vector is
  jointly down-sampled along time and channel dimensions.
  All backbones output the predicted mobility flow for the next $T$ hours.}
  \label{fig:pipeline}
\end{figure}
\subsection{Problem formulation}

Let $\mathbf{x}\!\in\!\mathbb{R}^{N\times M}$ be the multivariate time
series containing $N$ time steps (hours) and $M$ features (H3 cells).
For a look-back window of length $L$ and a forecasting horizon $T$, the
task is
\[
\bigl(\hat{\mathbf{x}}_{t+1},\dots,\hat{\mathbf{x}}_{t+T}\bigr)
      = f_\theta\!\bigl(\mathbf{x}_{t-L+1},\dots,\mathbf{x}_{t}\bigr),
\qquad t\in[L,N-T],
\]
where $f_\theta$ denotes a learnable mapping.

\paragraph{SPN-enhanced input.}
Prior to modelling we apply Spatial Neighbourhood Fusion (SPN):
for each cell $c$ and hour $t$ we compute the
neighbourhood median $\hat f_c^{(t)}$ and the mean
$\bar f_c^{(t)}=\tfrac12(f_c^{(t)}+\hat f_c^{(t)})$ of the original flow
$f_c^{(t)}$.  The model therefore ingests a three-channel tensor
\([\;f_c^{(t)},\hat f_c^{(t)},\bar f_c^{(t)}]\).

\subsection{Backbone library}
We evaluate five representative time-series architectures in this work: \textbf{NLinear} \cite{Zeng_AreTE_2022}, a parameter-efficient baseline built from stacked channel-wise linear layers; \textbf{LSTM} \cite{hochreiter1997long} and \textbf{GRU} \cite{chung2014empirical}, two gated recurrent models that process the sequence sequentially to capture long-range dependencies, implemented as stacked layers over time (either per channel or over concatenated multivariate input) with per-layer complexity $O(L)$; \textbf{PatchTST} \cite{Nie-2023-PatchTST}, which partitions the series into patches of length $L_p{=}4$ and feeds them into a Transformer encoder with overall complexity $O(L/L_p)$; and \textbf{Kernel U-Net (K-U-Net)} \cite{you_kun_2024}, a symmetric encoder–decoder whose blocks wrap custom kernels, using a Linear$\!\!\rightarrow$MLP$\!\!\rightarrow$Transformer hierarchy in the encoder and its mirror in the decoder, with skip connections preserving high-resolution context (see Fig.~\ref{fig:pipeline}).

\paragraph{\texttt{+spn} Variants.}
For each backbone, we train two versions: (i) a vanilla single-channel model, and (ii) an SPN-enhanced three-channel counterpart, denoted \textit{NLinear+spn}, \textit{PatchTST+spn}, \textit{K-U-Net+spn}, \textit{GRU+spn}, and \textit{LSTM+spn}. All hyperparameters are kept identical, except for the input dimensionality, so that any observed performance gain can be attributed solely to the spatial fusion mechanism.

\subsection{Dataset}
After filtering (Sec.~III) the mobility cube contains
$N{=}\,577$ hourly steps and $M{=}\,101$ H3 cells.
We adopt a chronological split of $70\%/10\%/20\%$
for training, validation, and testing.

\section{Experiments and Results}\label{sec:exp}

\subsection{Experimental settings}\label{ssec:setup}

Because public-health planners are interested in trends rather than
momentary peaks, we follow the pipeline in Fig.2 and predict the
SPN-smoothed flow series.  
For each H3 cell we form
$\smash{\textit{flow\_clean}_t=\alpha\,\textit{flow\_total}_t+
(1-\alpha)\,\widetilde{\textit{median}}_t}$,
where the tilde denotes the median of its six neighbours and  
$\alpha{=}0.5$ is chosen on the validation split (grid–search
$\alpha\!\in\!\{1.0,0.9,\dots,0.0\}$).

Six models—
\textit{NLinear}, \textit{NLinear+spn},\textit{GRU}, \textit{GRU+spn},\textit{LSTM}, \textit{LSTM+spn}, \textit{PatchTST},
\textit{PatchTST+spn}, \textit{K-U-Net}, and \textit{K-U-Net+spn}—
are benchmarked on look-back windows
$L\!\in\!\{48,72,96,120\}$ with a forecasting horizon of
$T\!=\!48$\,h.
Each “\,+spn” variant receives an extra channel that contains the
lag-1 neighbour–median; its backbone only sees the lag-1
self-history.  

\textbf{Optimisation.}
All runs minimise MSE with the Adam optimiser
(\textit{lr}$\,{=}5{\times}10^{-4}$),
batch size 128, and early stopping on a 10 \% validation split
(patience 50).
PatchTST and K-U-Net use a hidden size of 128;  
K-U-Net width factors are
$\!\langle6,2\rangle$, $\!\langle6,3\rangle$,
$\!\langle6,4\rangle$, $\!\langle6,5\rangle$
for $L\!=\!48,72,96,120$, respectively.
Each configuration is repeated with five random seeds and we report the
mean and—where space permits—its standard deviation.
\subsection{Main results}\label{ssec:main}
Table~\ref{tab:multivariate-table} reports train/test MSE on the smoothed target. Across all backbones, adding SPN features \textbf{consistently reduces error}. The largest relative drop is observed with \textbf{Linear+spn}, yielding average improvements of \textbf{46.13\%} on the train split and \textbf{50.49\%} on the test split. Among recurrent models, \textbf{GRU+spn} improves by \textbf{4.58\%} (train) and \textbf{14.70\%} (test), while \textbf{LSTM+spn} achieves mean gains of \textbf{4.11\%} and \textbf{9.23\%}, respectively. For Transformer-based methods, \textbf{PatchTST+spn} reaches \textbf{9.74\%} (train) and \textbf{8.00\%} (test) improvements. Finally, the convolutional \textbf{K-U-Net+spn} delivers strong gains of \textbf{9.92\%} (train) and \textbf{8.84\%} (test). These benefits hold across all look-back windows $L\!\in\!\{48,72,96,120\}$, with especially strong reductions for the Linear backbone and consistent improvements for recurrent, Transformer, and CNN models alike.

\section{Conclusion}
Leveraging nationwide mobility traces from Peru’s \textsc{DCT} application,
we proposed a lightweight Spatial-Neighbourhood Fusion (SPN) step and showed
that it consistently improves multiple forecasting backbones.
Average error reductions range from \textbf{8--10\%} on strong neural
backbones (PatchTST, K-U-Net, LSTM/GRU) up to \textbf{50\%} on the Linear
baseline with minimal overhead. Notably, the convolutional
\textit{K-U-Net} matches or surpasses \textit{PatchTST} in accuracy while
being roughly \textbf{2$\times$ faster} in training and inference, making it
well-suited for real-time deployment. These results highlight the value of
local spatial context and show that a simple, model-agnostic fusion step can
substantially enhance mobility forecasting in public-health settings.

\begin{appendices}





\section{Ethics Statement}

This study uses anonymized mobility data collected through Peru’s national Digital Contact Tracing (DCT) application, developed under CONCYTEC Project No. 70744. All personal identifiers were irreversibly hashed using SHA-256, and data collection was based on informed user consent. Participation was voluntary, and only aggregated, non-identifiable data were accessed by researchers. The study complies with all applicable data protection regulations and ethical standards.
\end{appendices}


\bibliography{sn-bibliography}

\end{document}